\RequirePackage{fix-cm}
\documentclass[twocolumn]{svjour3}          

\usepackage{booktabs}
\usepackage{multirow}

\usepackage{graphicx}
\usepackage{multirow}
\usepackage{multicol}
\usepackage{cite}
\usepackage{amsfonts}
\usepackage{amssymb}
\usepackage{rotating}  
\usepackage[cmex10]{amsmath}
\usepackage{caption}
\usepackage{colortbl}
\usepackage{subcaption}               
\usepackage{diagbox}
\usepackage{makecell}

\definecolor{Lightgray}{RGB}{235,235,235}
\usepackage[switch]{lineno} 
\begin{document}
\title{Microaneurysm Detection in Fundus Images Using a Two-step Convolutional Neural Networks}
\author{Noushin~Eftekheri, Mojtaba~Masoudi, Hamidreza~Pourreza, Kamaledin~Ghiasi Shirazi, Ehsan~Saeedi}
\institute{Department of Computer Engineering, Ferdowsi University of Mashhad,\\
              Mashhad, Iran\\
             \newline N. Eftekheri \\
             \email{Noushin.Eftekhari @stu.um. ac.ir}\\
              \and
              \newline M. Masoudi \\
              \email{m.masoudi @mail.um. ac.ir}\\
              \and
             \newline Dr. H. Pourreza\\
             \email{hpourreza@um.ac.ir}\\             
              \and
             \newline Dr K.Ghiasi Shirazi\\
             \email{ooo@um.ac.ir}\\             
             \and
            \newline Dr. E. Saeedi\\
            \email{saeedi.ehsan@gmail.com}\\
            }
\date{Received: date / Accepted: date}
\maketitle

\begin{abstract}
Diabetic Retinopathy (DR) is a prominent cause of blindness in the world. The early treatment of DR can be conducted from detection of microaneurysms (MAs) which appears as reddish spots in retinal images. An automated microaneurysm detection can be a helpful system for ophthalmologists. In this paper, deep learning, in particular convolutional neural network (CNN), is used as a powerful tool to efficiently detect MAs from fundus images. In our method a new technique is used to utilise a two-stage training process which results in an accurate detection, while decreasing computational complexity in comparison with previous works. To validate our proposed method, an experiment is conducted using Keras library to implement our proposed CNN on two standard publicly available datasets. Our results show a promising sensitivity value of about 0.8 at the average number of false positive per image greater than 6 which is a competitive value with the state-of-the-art approaches. 
	\keywords{Diabetic Retinopathy (DR), Microaneurysm (MA), Deep learning, Convolutional Neural Network (CNN).}
\end{abstract}
\section{Introduction}

Diabetic retinopathy (DR) is a diabetic disorder caused by changes in the blood vessels of the retina. The damage of retinal blood vessels may end in blindness. DR occurs in most people with diabetes and its treatment depends on the patient's age and the duration of DR. In 2000, WHO estimates that 171 million people have diabetes and 366 million cases will occur in 2030 \cite{imani2015fully}. DR can be treated effectively with laser therapy if detected early. The important symptoms of diabetes are swelling of the blood vessels, fluid leakage in eyes and also in some cases the growth of new blood vessels on the retina. 80\% of patients with diabetes have DR for more than 10 years\cite{zhang2011retinal}. Microaneurysm (MA) is the first symptom of DR that causes blood leakage to the retina. This lesion usually appears as small red circular spots with a diameter of fewer than 125 micrometers \cite{imani2015fully}. 

Although it is claimed that recently diabetic retinopathy can be detected directly, we cannot ignore the importance of microaneurysm detection as a field of research, in clinical decision making such as checking patient status, type and duration of treatment, and also its application in more specific treatments such as detection of pigmentary retinopathy. Only few papers directly address these issues and we expect that more emphasis will be given to those areas in the near future, for example in the application of multi-stream networks in a fully convolutional fashion.
As mentioned in \cite{agrawal2013survey}, the methods of automatic MA detection proceed in three stages (preprocessing, MA candidate extraction and classification). The preprocessing stage is to reduce the noise and enhance the contrast of input images by applying image preprocessing techniques. These techniques are performed on the green colour plane of RGB images, because in this plane microaneurysms have the higher contrast with the background. In the second stage (MA candidate extraction stage), candidate regions for MA are detected, and because many of the blood vessels may result in false positives, the blood vessels are extracted from the candidates using blood vessel segmentation algorithms. In the third stage (classification stage), after applying feature extraction and selection, a classification algorithm is used to categorize features into MA candidate (abnormal) and non-MA candidate (normal), while a probability is estimated for each candidate using a classifier and a large set of specifically designed features to represent a MA.

Over the past decade, there has been a dramatic increase in the research about automatic microaneurysms (MA) detection. First MA detection was introduced based on mathematical morphology detection approach in fluorescein angiograms. In these methods, to  distinguish MA from vessels a morphological top-hat transformation with a linear structuring element at different orientations is used \cite{spencer1996image}. Then, some papers, such as Spencer, Cree, Frame, and coworkers \cite{spencer1992automated,frame1998comparison}, tried to improve these approaches. They added two more steps to the basic top-hat transform based detection technique, matched filtering postprocessing step and a shade-correction preprocessing step. After detection and segmentation of candidate MA, various shape and intensity based features were extracted and finally a classifier was used to separate the real MA from spurious responses. After that, a modification version of the top-hat based algorithm is proposed and applied to high-resolution, red free fundus photographs by Hipwell et al. \cite{hipwell2000automated}. Fleming et al. \cite{fleming2006automated} proposed an improvement of this method by locally normalizing the contrast around candidate lesions and eliminating candidates detected on vessels through a local vessel segmentation step. Niemeijer et al. \cite{niemeijer2005automatic} detect MA candidates in color fundus images through presenting a hybrid scheme that used both the top-hat based method as well as a supervised pixel classification based method.

In addition to mathematical morphology approach, some other techniques are used for detection of red lesion in fundus image. Sinthanayothin et al. \cite{sinthanayothin2002automated} assumed that images can be categorized in three classes, vessels, red lesion and MA. Then after applying neural networks as a recursive region growing procedure to segment both the vessels and red lesions in a fundus image, any remaining objects were identified as microaneurysms. 
Kamel et al. \cite{kamel2001neural} also utilized a NN approach for automatic detection of MA in retinal angiograms. Using NN provide the ability of detecting the regions with MA and rejecting other regions. Usher et al. \cite{usher2004automated} used neural network to detect the microaneurysms. First of all preprocessing is done. After preprocessing, microaneurysms are extracted using recursive region growing and adaptive intensity thresholding with “moat operator” and edge enhancement operator. Moreover, Quellec et al.\cite{quellec2008optimal} described a supervised MA detection method based on template matching in wavelet-subbands.
%


However, literature reviews have indicated that there are still some problems in this area that haven't been considered before \cite{zhang2011retinal}. One of the main problems in MA detection is the poor quality images in JPEG format of the publicly available datasets, so MAs are too blurred or too small to be detected. Moreover, considering that the scale of Gaussian kernel is fixed, different size of MAs cannot be covered properly. Larger MA will not be extracted if only small scale is used, and if a large scale is used, then small MAs that lie close to each other are considered as one MA. This, produces a lower correlation coefficient. Furthermore, a few MAs which are located close to blood vessels are missed in preprocessing stage, because they are recognized as part of the vascular map which should be removed in this stage. Another challenge of MA detection using neural networks, is imbalanced dataset which means the number of Non-MA samples are usually much higher than the number of MA ones. This can lead to improper and imbalanced training of networks, and also decrease classification accuracy.     

In this paper, a new method for MAs detection in fundus images based on deep-learning neural networks is developed to address the problems with the current automatic detection algorithms. Only few papers directly address issues specific to object detection like class imbalance. Deep learning algorithms, in particular convolutional networks, have rapidly become a methodology of choice for analyzing medical images \cite{lecun2015deep}. Deep learning is an improvement of artificial neural networks, consisting of more layers that permit higher levels of abstraction and improved predictions from data \cite{gu2015recent}. In our proposed method, by using the characteristic of convolution neural networks, the MA candidates are selected from the informative part of the image where their structure is similar to an MA, then a CNN will detect the MA and Non-MA spots. 
According to our results, the proposed method can decrease false-positive rate and can be considered as a powerful solution for automatic MA-detection approach. 

This paper starts with a brief introduction to deep learning in Part 2. Part 3 is dedicated to our proposed method. Then Part 4 shows our experimental results, and Part 5 is dedicated for discussion. Finally, in part 6, we conclude our work.

\section{Deep Learning in Medical Image Analysis}
Artificial neural networks and deep learning, conceptually and structurally inspired by neural systems, rapidly become an interesting and promising methodology for researchers in various fields including medical imaging analysis. Deep learning means learning of the representations of data with multiple levels of abstraction used for computational models that are composed of multiple processing layers. These methods rapidly become an interesting and promising methodology for researcher and are gaining acceptance for numerous practical applications in engineering. Deep learning has performed especially well as classifiers for image-processing applications and as function estimators for both linear and non-linear applications. Deep learning recognize complicated structure in big data sets by utilizing the back propagation algorithm to indicate how the internal parameters of a NN should be changed to compute the representation in each layer from the representation in the previous layer \cite{lecun2015deep}.

In particular, convolutional neural networks (CNNs) automatically learn mid-level and high-level abstractions obtained from raw data (e.g., images), and so have been considered as powerful tools for a broad range of computer vision tasks \cite{lecun2015deep}. Recent results indicate that the generic descriptors extracted from CNNs are extremely effective in object recognition and localization in natural images \cite{lecun2015deep}. Medical image analysis are quickly entering the field and applying CNNs and other deep learning methodologies to a wide variety of applications \cite{lecun2015deep,van2016fast}. In medical imaging, the accurate diagnosis of a disease depends on both image acquisition and image interpretation. Thanks to the emerging of modern devices acquiring images very fast and with high resolution, image acquisition has improved substantially over recent years. The image interpretation process, however, has just recently begun to benefit from machine learning.
%
%
%

\subsection{Convolutional Neural Networks (CNNs)}
Convolutional neural networks (CNNs) is one of the successful type of models for pattern recognition and classification in image analysis. 
CNN consists of a set of layers called convolutional layers each of which contains one or more planes as a feature map. Each unit in a plane receives input from a small neighborhood in the planes of the previous layer. Each plane  has a fixed feature detector that is convolved with a local window which is scanned over the planes in the previous layer to detect increasingly more relevant image features, for example lines or circles that may represent straight edges or circles, and then higher order features like local and global shape and texture. To detect multiple features, multiple planes are usually used in each layer. The output of the CNN is typically one or more probabilities or class labels \cite{van2016fast}.



\begin{figure} 
	\centering
	\includegraphics[scale=0.37]{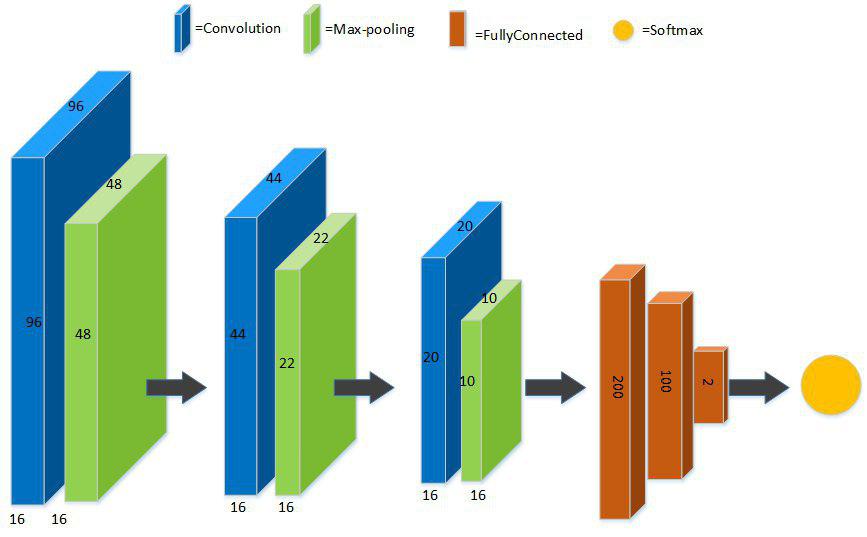}
	\caption{The architecture of basic CNN applied in this project.}
	\label{1}
\end{figure}

Fig.\ref{1} shows one of the architecture of CNN structured we used in MA detection. As can be seen, the network is designed as a series of stages. The first three stages are composed of convolutional layers (blue) and pooling layers (green) and the output layer (brown) is consist of three fully-connected layers and the last layer is the softmax function. The details are explain in section 4.2. 
\section{The proposed method}
\begin{figure*}
	\centering
	\includegraphics[scale=0.6]{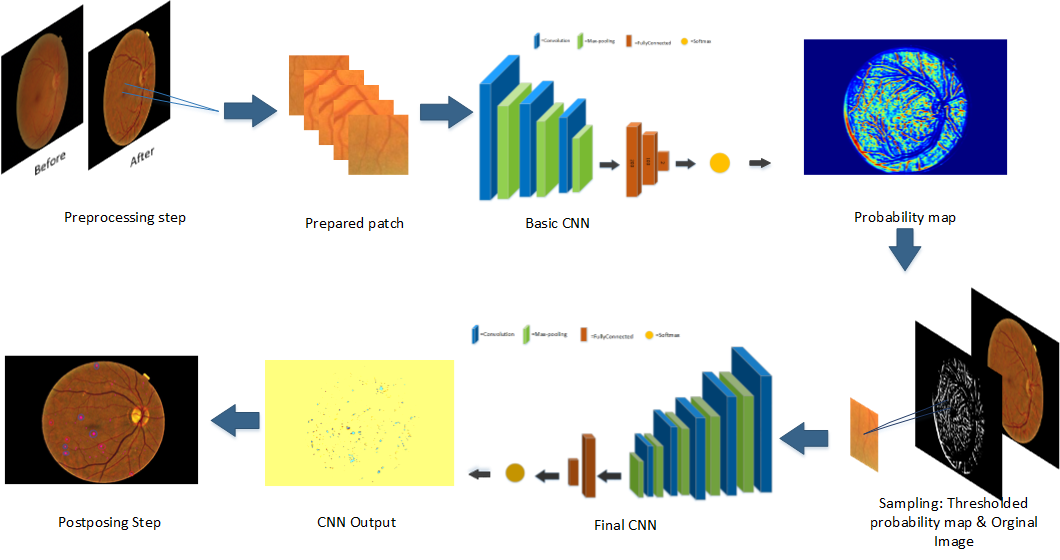}
	\caption{Illustration of the proposed method.}
	\label{2}
\end{figure*}

Only few papers
directly address issues specific to object detection
like class imbalance/hard-negative mining or efficient
pixel/voxel-wise processing of images. We expect that
more emphasis will be given to those areas in the near
future, for example in the application of multi-stream
networks in a fully convolutional fashion

To address the usual problems of previous works, mentioned in Introduction (poor quality of images, the fixed scale of Gaussian kernel, MAs located close to blood vessels and imbalanced dataset), we proposed a two-stage training strategy where informative normal samples are selected from a probability map which is the output of the first CNN, called basic CNN. The final CNN classify each pixel in the test images as MA or non-MA. This CNN gets the probability map from the previous stage as the selected samples for the input test images, and result in a final probability-map for each test image showing the probability of being a pixel MA or non-MA. Fig.\ref{2} shows different steps of the proposed method.

\subsection{Preprocessing Step}
Because the retinal images are usually non-uniform illumination, a preprocessing step is needed to apply colour normalization and eliminate retina background. This procedure was accomplished by estimating the background image and subtracting that from the original image. The background image was obtained by median filtering the original image with a 30*30 pixel kernel. 

Afterwards, input patches with the size of $101\times101$ are produced for training of the basic CNN. These patches are labeled based on the label of their central pixel from ground truth dataset. Those with a MA pixel at the center are considered as MA samples and those with non-MA pixel are considered as non-MA samples for training.

\subsection{Candidate Selection by basic CNN}
In this stage the basic CNN is trained with small patches whose labels are determined by the label of their central pixel. The basic CNN is used to solve the imbalanced data problem where the number of patches including MA and Non-MA in retinal images is not balanced and cause network complexity and improper convergent, because using a deep complex network rather than a 2-stage simple network, is harder to train and lots of uninformative data are involved. To avoid imbalanced data problem, in each epochs, an equal number of MA and non-MA patches are selected to train the network. However, because of not using all non-MAs in learning process, selecting the equal number of MA and non-MA patches will cause a high false positive in initial results. The basic CNN returns an initial probability map indicating for each input pixel the initial probability of belonging to MA. This map is needed to prepare the input dataset for the final CNN. Fig.\ref{1} shows the architecture of basic CNN. The training procedure in CNN is a sequential process that requires multiple iterations to optimize the parameters and extract distinguishing characteristics from images. In each iteration, a subset of samples are chosen randomly and applied to optimize the parameters. This is obtained by back propagation (BP) and minimizing cost function \cite{lecun2015deep}. 

\subsection{Classification by final CNN}
The final CNN works as the main classifier to extract the MA candidate regions. This CNN has more layers, and therefore more abstract levels than the basic CNN which lead to a discriminative MA modelling. Unlike the basic CNN which used a random sample from the input dataset pool, the final CNN apply the probability map from the previous stage as the selected samples for the input images. In other words, the CNN inputs are selected from the original image samples which are chosen by the corresponding pixels of a thresholded probability map (Figure \ref{1}). This map decides which probabilities can be considered as MAs or non-MAs candidate. Therefore, the CNN inputs would be informative patches whose structure is similar to MA. 
The output of this CNN is a map for each test image showing the MA probability of a pixel. However, this map is noisy and a post-processing step is needed. 
\subsection{Post-processing} 

In practice, the probability map obtained from the final CNN is extremely noisy. For example when there is two close candidates, they are merged and considered as one. Therefore, to obtained a smoothed probability map, it is convolved with a 5-pixel-radius disk kernel. The local maximum of the new map are expected to lie at the disk centres in the noisy map, i.e., at the centroids of each MA to obtain a set of candidates for each image.
%
%
%

\subsection{The architectures of CNNs}
In this work, two different structure are used for the basic and final CNNs. As can be seen from Fig.\ref{1}, the basic CNN includes three convolution layers, each of which followed by a pooling layer, then  three fully-connected layer and finally a Softmax layer in the output layer. The final CNN has more layers than the basic CNN. The corresponding layer number of final CNN is five convolution and pooling layers, then two fully-connected and one Softmax classification layer which is fully connected with two neurons for MA and non-MA, see Table \ref{t1} and \ref{t2}.

%
%
%
\begin{table*}[]
	\centering
	\caption{Architectures of Final CNN}
	\label{t1}
	\begin{tabular}{|l|l|c|l|c|}
		\hline
		Layer   & Operation                          & \multicolumn{1}{l|}{Input size} & Detail                 & Berr,(p) \\ \hline
		Layer1  & \multicolumn{1}{c|}{Input}         & $3\times101\times 101$                        & \multicolumn{1}{c|}{-} & -        \\ \hline
		Layer2  & \multicolumn{1}{c|}{Convolutional} & $16\times101\times 101$                      & $6\times6$                    & -        \\ \hline
		Layer3  & \multicolumn{1}{c|}{Max Pooling}   & $16\times50\times 50$                        & $2\times2$                   & 0.25     \\ \hline
		Layer4  & Convolutional                      &  $16\times48\times 48$                        & $5\times5$                    & -        \\ \hline
		Layer5  & Max Pooling                        &  $16\times24\times 24$                        & $2\times2$                    & -        \\ \hline
		Layer6  & Convolutional                      & $16\times22\times 22$                         & $3\times3$                    & -        \\ \hline
		Layer7  & Max Pooling                        & $16\times11\times 11$                          & $2\times2$                    & 0.25     \\ \hline
		Layer8  & Convolutional                      & $16\times10\times 10$                          & $2\times2$                    & -        \\ \hline
		Layer9  & Max Pooling                        & $16\times5\times5$                         & $2\times2$                    & -        \\ \hline
		Layer10 & Convolutional                      & $16\times4\times 4$                           & $2\times2$                    & -        \\ \hline
		Layer11 & Max Pooling                        & $16\times2\times 2$                           & $2\times2$                    & -        \\ \hline
		Layer12 & Fully Connected                    & 100                             & $1\times1$                    & -        \\ \hline
		Layer13 & Fully Connected                    & 2                               & $1\times1$                    & -        \\ \hline
	\end{tabular}
\end{table*}

\begin{table*}[]
	\centering
	\caption{Architectures of Basic CNN}
	\label{t2}
	\begin{tabular}{|l|l|c|l|c|}
		\hline
		Layer   & Operation                          & \multicolumn{1}{l|}{Input size} & Detail                 & Berr,(p) \\ \hline
		Layer1  & \multicolumn{1}{c|}{Input}         & $3\times101\times 101$                      & \multicolumn{1}{c|}{-} & -        \\ \hline
		Layer2  & \multicolumn{1}{c|}{Convolutional} & $16\times96\times 96$                       & $6\times6$                    & -        \\ \hline
		Layer3  & \multicolumn{1}{c|}{Max Pooling}   & $16\times48\times 48$                        & $2\times2$                     & 0.25     \\ \hline
		Layer4  & Convolutional                      & $16\times44\times 44$                        & $5\times5$                     & -        \\ \hline
		Layer5  & Max Pooling                        & $16\times22\times 22$                        & $2\times2$                    & 0.25     \\ \hline
		Layer6  & Convolutional                      & $16\times20\times 20$                         & $3\times3$                    & -        \\ \hline
		Layer7  & Max Pooling                        & $16\times10\times 10$                      & $2\times2$                     & 0.25     \\ \hline
		Layer8  & Fully Connected                    & 200                             & $1\times1$                     & -        \\ \hline
		Layer9  & Fully Connected                    & 100                             & $1\times1$                   & -        \\ \hline
		Layer10 & Fully Connected                    & 2                               & $1\times1$                   & -        \\ \hline
	\end{tabular}
\end{table*}

In this work, to increase the accuracy, a dropout training with a maxout activation function is used. Dropout means to reduce over-fitting by randomly omitting the output of each hidden neuron with a probability of 0.25. 

Training process is similar to standard neural network using stochastic gradient descent. We have incorporated dropout training algorithm for three convolutional layers and one fully connected hidden layer. 16 filter sizes $6\times 6$ in the first convolution layer, 16 filter size $5\times 5$ in the second layer, and 16 filter size $3\times 3$ is applied in the third convolution layer, and then maxout activation function is used for all layers in the network except for the softmax layer.The filter size in Max pool layer is $2\times 2$ with stride 2. After each pair convolution and pooling layers, an activation LeakyReLU layer is applied that improved the version of ReLU (rectify linear unit). In this version, unlike the ReLU in which negative values become zero and so neurons become deactivated, these values in the Leaky ReLU will not be zero, instead, the value of $a$ is added to the Equation \ref{eq2}.

\begin{equation} \label{eq2}
f(x)=
\begin{cases}
x       & \quad \text{ x}\geq0\\
a x  & \quad  \text{ otherwise}
\end{cases}
\end{equation}

Where $a$ is a small constant value (0.01) and $x$ is the output of the previous layer. The final layers of the network consist of a fully connected layer and a final Softmax classification layer. This function produces a score ranging between $0$ and $1$, indicating the probability of pixel belongs to the MA class. To train the network, loss function of a binary cross entropy is used, note that for a 2 class system output $t2=1 - t1$. cross entropy calculate the difference between predicted values (p) and targets (t), using the following equation:

\begin{equation} \label{eq4}
L = -t\log(p) - (1-t)\log(1-p)
\end{equation}
\section{Experimental Results}
To verify our proposed method, we implement the CNNs using deep learning Keras libraries based on Linux Mint operating system with 32G RAM, Intel (R) Core (TM) i7-6700K CPU and  NVIDIA GeForce GTX 1070 graphics card. 
In this experiment, we used  two standard publicly available datasets, ROC \cite{rocdataset} and E-Ophtha-MA \cite{ophtadataset} databases to train and test the proposed method for the detection of MA in retinal images. ROC includes 100 colour image of the retina that obtained from Topcon NW 100, Topcon NW 200 and Canon CR5-45NM cameras with JPEG format. These images were divided into two parts of 50 subsets of training and testing. The image dimensions are $768\times 576$ , $1058\times 1061$ and $1389\times 1383$  \cite{niemeijer2010retinopathy}. E-Ophtha-MA database contains 148 colour images of JPEG format and with the size of $2544 \times1696$. Totally there are 248 images in our database and many patches are extracted from each image. For our training and testing inputs we used about  28786 MA + 258354 NonMA patches. Moreover, data augmentation is used by mirroring and rotating patches.

To validate results, a cross-validation algorithm is utilized to divide the data to 75\% training and 25\% testing sets, then exchange the training and testing sets in successive rounds such that all data has a chance of being trained and tested. For accuracy evaluation, we computed true positive (TP) as the number of MA pixels correctly detected, false positive (FP) as the number of non-MA pixels which are detected wrongly as MA pixels, in other words detected pixels which had no reference of MA within a 5-pixel-radius of our disk kernel, false negative (FN) as the number of MA pixels that were not detected and true negative (TN) as the number of no MA pixels which were correctly identified as non-MA pixels. For better representation of accuracy, sensitivity is defined as follow.

\begin{equation} \label{eq3}
sensitivity=\frac{TP}{TP+FN}
\end{equation}
In this experiment, to verify the accuracy of the proposed method, we compared our sensitivity value  with the usual current works (Latim\cite{quellec2008optimal}, OkMedical \cite{zhang2010detection}, Waikato \cite{cree2008waikato}, Fujita Lab\cite{hatanaka2012automated}, B Wu's method \cite{wu2017automatic}, Valladolid \cite{sanchez2009mixture}), on E-Ophtha-MA data set in Table \ref{t4} and ROC dataset in Table \ref{t5}.

In addition, to assess our result, ROC evaluation algorithm[2010 Retinopathy…] is applied and the output of this algorithm is then used to generate an FROC curve [24] that plots the sensitivity against the average number of false positive detection per image Fig.\ref{34}. 

Moreover, Table \ref{t3} shows Competition performance measure (CPM) to evaluate our results. This table confirm the success of the proposed method. 
\section{ِِِDiscussion}
  
\definecolor{Lightgray}{RGB}{235,235,235}
\begin{table*}[]
	\centering
	\caption{Sensitivities of the different methods in ROC dataset at the various FP/image rates.}
	\label{t4}
	\begin{tabular}{|c|l|l|l|l|l|l|l|l|}
		\hline
		\multicolumn{9}{|c|}{\begin{tabular}[c]{@{}c@{}}
				\rowcolor{Lightgray}FROC results on Retinopathy online challenge dataset at\\  average number of False positives per image\end{tabular}} \\ \hline
		\multirow{7}{*}{\begin{sideways} Sensitivity~ \end{sideways}}         &\theadfont\diagbox[width=7em]{Method}{FPs/Img}     & 1/8         & 1/4         & 1/2         & 1           & 2           & 4           & 8           \\ \cline{2-9} 
		& Proposed              & 0.04        & 0.17        & 0.35        & 0.55        & 0.61        & 0.72        & 0.76        \\ \cline{2-9} 
		& Valladolid  \cite{sanchez2009mixture}         & 0.19        & 0.22        & 0.25        & 0.30        & 0.36        & 0.41        & 0.52        \\ \cline{2-9} 
		& Waikato\cite{cree2008waikato}             & 0.06        & 0.11        & 0.18        & 0.21        & 0.25        & 0.30        & 0.33        \\ \cline{2-9} 
		& Latim\cite{quellec2008optimal}               & 0.17        & 0.23        & 0.32        & 0.38        & 0.43        & 0.53        & 0.60        \\ \cline{2-9} 
		& OkMedical\cite{zhang2010detection}           & 0.20        & 0.27        & 0.31        & 0.36        & 0.39        & 0.47        & 0.50        \\ \cline{2-9} 
		& Fujita Lab\cite{hatanaka2012automated}        & 0.18        & 0.22        & 0.26        & 0.29        & 0.35        & 0.40        & 0.47        \\ \hline
	\end{tabular}
\end{table*}
\begin{table*}[]
	\centering
	\caption{Sensitivities of the different methods in E-Ophtha-MA dataset at the various FP/image rates.}
	\label{t5}
	\begin{tabular}{|c|l|l|l|l|l|l|l|l|}
		\hline
		\multicolumn{9}{|c|}{\begin{tabular}[c]{@{}c@{}}
				\rowcolor{Lightgray}FROC results on Retinopathy online challenge dataset at\\  average number of False positives per image\end{tabular}} \\ \hline
		\multirow{3}{*}{\begin{sideways} \scriptsize{Sensitivity~} \end{sideways}}        &\theadfont\diagbox[width=6em]{Method}{~ ~FPs/Img}   & 1/8          & 1/4          & 1/2          & 1            & 2            & 4            & 8            \\ \cline{2-9} 
		& Proposed        & 0.09         & 0.26         & 0.40         & 0.53         & 0.58         & 0.67         & 0.77         \\ \cline{2-9} 
		& B Wu's\cite{wu2017automatic}          & 0.06        & 0.12        & 0.17        & 0.24        & 0.32        & 0.42        & 0.57        \\ \hline
		
	\end{tabular}
\end{table*}
 
\begin{figure}[]
	\centering

	\begin{subfigure}[b]{0.55\textwidth}
		\includegraphics[width=\textwidth]{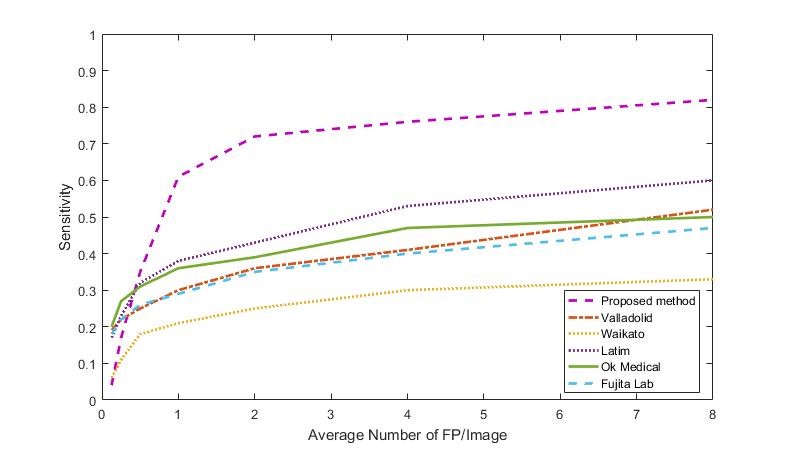}
		\caption{FROC curves for ROC dataset}
		\label{3}
	\end{subfigure}
	\begin{subfigure}[b]{0.55\textwidth}
		\includegraphics[width=\textwidth]{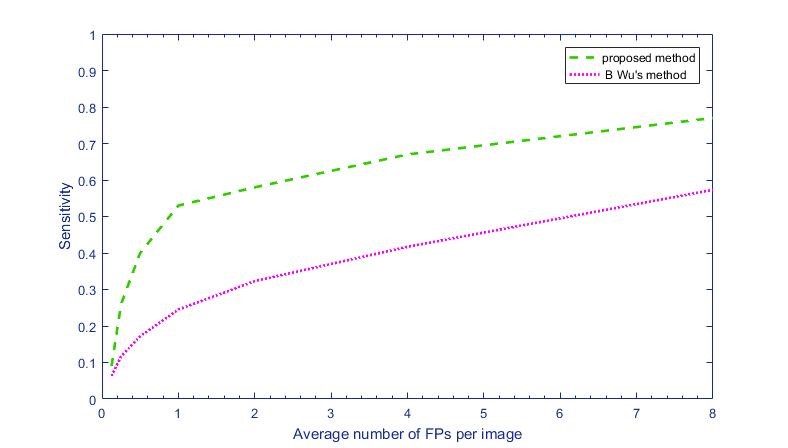}
		\caption{FROC curves for E-Ophtha-MA dataset}
		\label{4}
	\end{subfigure} 
	\caption{The comparison of FROC curves of the proposed and previous methods}
	\label{34}
\end{figure}

From Table \ref{t4} and \ref{t5}, our proposed method, compared with other methods, has the lowest sensitivity (0.04) when the average number of FP per image (FPs/Img) is $1/8$, while this values increased quickly and increased to a maximum of 0.76 at FPs/Img equals 8. Valladolid assume all pixels in the image are part of one of three classes: class 1 (background elements), class 2 (foreground elements, such as vessels, optic disk and lesions), and class 3 (outliers). A three class Gaussian mixture model is fit to the image intensities and a group of MA candidates are segmented by thresholding the fitted model.
The sensitivity of this method is 0.19 at FPs/Img $=1/8$ and gradually increase to 0.52 at FPs/Img $=8$.
The Waikato Microaneurysm Detector performs a top-hat transform by morphological reconstruction using an elongated structuring element at different orientations which detects the vasculature. After removal of the vasculature and a microaneurysm matched filtering step the candidate positions are found using thresholding. 
In comparison with other methods, Waikato has the lowest sensitivity ranging from 0.06 to 0.33.
Latim assumes that microaneurysms at a particular scale can be modelled with 2-D, rotation-symmetric generalized Gaussian functions. It then uses template matching in the wavelet domain to find the MA candidates. 
Latim method can be considered to have the second high sensitivity value after our proposed method. The sensitivity of this method is 0.17 at FPs/Img $=1/8$ and 0.60 at FPs/Img $=8$.
OkMedical responses from a Gaussian filter-bank are used to construct probabilistic models of an object and its surroundings. By matching the filter-bank outputs in a new image with the constructed (trained) models a correlation measure is obtained. 
In Fujita lab work, a double ring filter was designed to detect areas in the image in which the average pixel value is lower than the average pixel value in the area surrounding it. Instead, the modified filter detects areas where the average pixel value in the surrounding area is lower by a certain fraction of the number of pixels under the filter in order to reduce false positive detections on small capillaries. 
The sensitivity of OkMedical and Fujita ranged from 0.18 to 0.50.
Fig.\ref{34} confirm our results on Table \ref{t4} and \ref{t5}. This figure shows the free-response receiver operating characteristic (FROC) which is a graphical plot illustrating the diagnostic ability of classifiers and created by plotting the sensitivity (TP rate) against the average of FP per image at various threshold settings, and compare the sensitivity of the proposed method and other methods from \cite{quellec2008optimal,sanchez2009mixture,cree2008waikato,zhang2010detection,hatanaka2012automated} on ROC and E-Ophtha-MA databases. 
From Fig.\ref{3} we can see that the sensitivity of the proposed method on ROC dataset is about 0.3 higher that other methods. It is about 0.6 for the FP greater than 1 and reached the maximum of 0.8, while this number for other methods doesn't exceed 0.6. Fig.\ref{4} also shows that the sensitivity of the proposed methods on E-Ophtha-MA databse is about 0.2 greater than BWu's method\cite{wu2017automatic}.

In addition, Table \ref{t3}  also confirm that our proposed method has the highest Competition performance measure (CPM) value for both ROC and E-Ophtha-MA datasets. This value are 0.45 for ROC dataset, while the corresponding values for Antal and Lazar methods are 0.43 and 0.42 respectively and other methods are less than 0.38. The CPM value of our method for E-Ophtha-MA dataset is 0.42 in comparison with 0.35 of BWu's method.
\begin{table}[h]
	\centering
	\caption{Final Score(CPM)}
	\label{t3}
	\begin{tabular}{|l|c|c|}
		\hline
		\multicolumn{3}{|c|}{\begin{tabular}[c]{@{}c@{}}
				\rowcolor{Lightgray}Competition performance measure(CPM) of\\ retinopathy online challenge at different operating points\end{tabular}} \\ \hline
		Final Score                                                  & ROC                                          & E\_Ophtha\_MA                                         \\ \hline
		Propose-method                                               & 0.45                                         & 0.42                                                  \\ \hline
		
		Antal\cite{antal2012ensemble}                                         &0.43                                            & -        
		\\ \hline 
		
		Lazar \cite{lazar2013retinal}                                                & 0.42                                            & -        
		\\ \hline     
		
		Valladolid \cite{sanchez2009mixture}                                         & 0.32                                         &                                                       \\ \hline
		Waikato \cite{cree2008waikato}                                               & 0.21                                         & -                                                     \\ \hline
		Latim \cite{quellec2008optimal}                                               & 0.38                                         & -                                                     \\ \hline
		OkMedical \cite{zhang2010detection}                                            & 0.36                                         & -                                                     \\ \hline
		Fujita Lab\cite{hatanaka2012automated}                                          & 0.31                                         & -                                                     \\ \hline	B Wu's method \cite{wu2017automatic}                                                & -                                            & 0.35                                                  \\ \hline
                                    
	\end{tabular}
\end{table}
%



\begin{figure*}[]
	\centering
	\begin{subfigure}[b]{0.19\textwidth}
		\includegraphics[width=\textwidth]{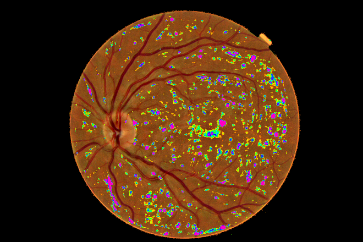}
	\end{subfigure}
	\begin{subfigure}[b]{0.19\textwidth}
		\includegraphics[width=\textwidth]{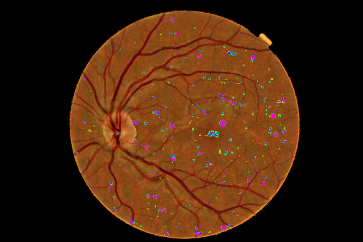}
	\end{subfigure} 
	\begin{subfigure}[b]{0.19\textwidth}
		\includegraphics[width=\textwidth]{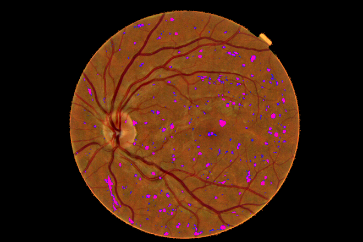}
	\end{subfigure}
	\begin{subfigure}[b]{0.19\textwidth}
		\includegraphics[width=\textwidth]{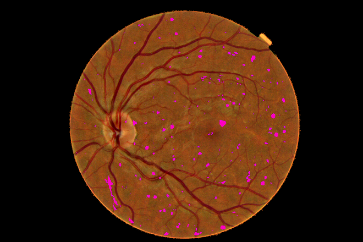}
	\end{subfigure}
	
	\begin{subfigure}[b]{0.19\textwidth}
		\includegraphics[width=\textwidth]{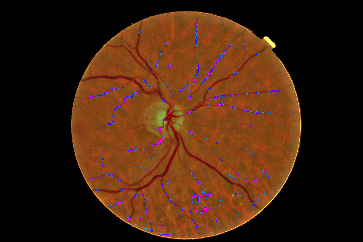}
	\end{subfigure}
	\begin{subfigure}[b]{0.19\textwidth}
		\includegraphics[width=\textwidth]{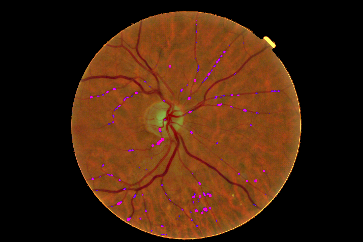}
	\end{subfigure} 
	\begin{subfigure}[b]{0.19\textwidth}
		\includegraphics[width=\textwidth]{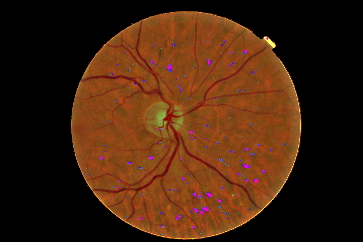}
	\end{subfigure}
	\begin{subfigure}[b]{0.19\textwidth}
		\includegraphics[width=\textwidth]{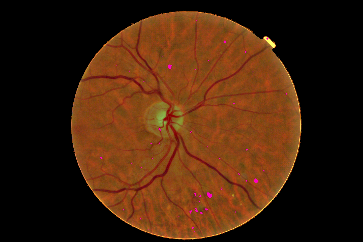}
	\end{subfigure}

	\caption{Pixel probability maps obtained from the final CNN for a different number of epochs. In initial epochs, the probability map include low probabilities of MA (depicted as green spots), in the subsequent epochs, the medium and high probabilities are in blue and purple respectively.}
	\label{77}
\end{figure*}



\section{Conclusion}  
In this paper, an approach for automatic MAs detection in retinal images based on deep-learning CNN is developed to address the previous works problems such as imbalanced dataset and inaccurate MA detection. In this method, because of using a two-stage CNN, the MAs candidate for classification process are selected from a balanced dataset and informative part of the image where their structure is similar to MA, and this results in decreasing computational complexity. According to our experimental results based on two standard publicly available dataset, the proposed method is about 0.3 higher than other methods. It has a promising sensitivity value of about 0.8 at the average number of false positive per image greater than 6 and can decrease false-positive rate compared to previous methods; it ,therefore, can be considered as a powerful improvement for previous MA-detection based on retinal images approach. For the future work, we plan to improve the training phase by combining the base and the final CNNs, and also apply this method on other medical application where imbalance data is an issue.
%
%
%

%
\section{Compliance with Ethical Standards}
\begin{itemize}
	\item \textbf{conflict of interest:} The authors (Noushin Eftekheri, Dr.Hamidreza Pourreza, Dr K.Ghiasi Shirazi and Dr.Ehsan Saeedi) declare that they have no conflict of interest.
	\item \textbf{Ethical Standards:} This article does not contain any studies with human or animal subjects performed by any of the authors; hence, formal consent is not applicable. In this study, two standard publicly available databases, ROC \cite{rocdataset} and E-Ophtha-MA \cite{ophtadataset} databases are used.
\end{itemize}

\end{document}